# Automated Processing of eXplainable Artificial Intelligence Outputs in Deep Learning Models for Fault Diagnostics of Large Infrastructures

G. Floreale[1], P. Baraldi[1], E. Zio[1,2], O. Fink [3]

[1]*Energy Department, Politecnico di Milano, Milano, Italy*
[2]*MINES Paris-PSL, Centre de Recherche sur les Risques et les Crises (CRC), Sophia Antipolis, France*
[3] *Intelligent Maintenance and Operations Systems (IMOS), EPFL, Lausanne, Switzerland*

## Abstract

Deep Learning (DL) models processing images to recognize the health state of large infrastructure components can exhibit biases and rely on non-causal shortcuts. eXplainable Artificial Intelligence (XAI) can address these issues but manually analyzing explanations generated by XAI techniques is time-consuming and prone to errors. This work proposes a novel framework that combines post-hoc explanations with semi-supervised learning to automatically identify anomalous explanations that deviate from those of correctly classified images and may therefore indicate model abnormal behaviors. This significantly reduces the workload for maintenance decision-makers, who only need to manually reclassify images flagged as having anomalous explanations. The proposed framework is applied to drone-collected images of insulator shells for power grid infrastructure monitoring, considering two different Convolutional Neural Networks (CNNs), GradCAM explanations and Deep Semi-Supervised Anomaly Detection. The average classification accuracy on two faulty classes is improved by 8% and maintenance operators are required to manually reclassify only 15% of the images. We compare the proposed framework with a state-of-the-art approach based on the faithfulness metric: the experimental results obtained demonstrate that the proposed framework consistently achieves $F_1$ scores larger than those of the faithfulness-based approach. Additionally, the proposed framework successfully identifies correct classifications that result from non-causal shortcuts, such as the presence of ID tags printed on insulator shells.

## 1. Introduction

Monitoring large infrastructures is important to ensure their reliable, cost-effective, safe and environmentally sustainable operation [1]. The main challenges include the complexity and large scale of the infrastructures, necessitating automated (big) data collection and analysis [2]. Recent progresses in the capability of collecting (e.g. via flying/walking robots and underwater vehicles) and storing a large number of high-resolution images [3], along with advancements in data processing technologies, have enabled the monitoring of critical components of large infrastructures (e.g. insulators, transformers and cables of the electrical system) using images [4]. The aim is to support the planning of timely maintenance interventions to prevent damages or accidents [5]. In particular, Deep Learning (DL) models are used to classify images to assess the health state of components [6]. However, if the training dataset is imbalanced, contains biases, or unrepresentative images, the DL models can learn and perpetuate non-general relations, reducing their accuracy and trustworthiness. Also, while DL models excel at finding correlations within the data, they may rely on non-causal shortcuts, i.e. spurious correlations that do not reflect the true causal relationships [7]. The consequence is that DL models can provide accurate classifications using features that are irrelevant or not correlated to the component's health state. Therefore, it is necessary to ensure that the process employed by DL models to generate outputs from inputs is based on causal relationships [2].

Table 1: Symbols used throughout the manuscript

| **Main Symbols** |
|---|
| $x$: generic image used to test the framework |
| $\Phi$: generic classification model; $\Phi^E$: EfficientNet B-0 classifier; $\Phi^M$: MObileNetV3 Small classifier |
| $Z$: number of classes |
| $\hat{z}$: predicted class |
| $g$: model providing the explanations |
| $y$: explanation |
| $\Psi$: model which identifies if an explanation is normal or anomalous |
| $\hat{l}$: binary label indicating if the explanation is identified as normal (+1) or anomalous (-1) |
| $D$: dataset of images used to develop the framework |
| $T$: number of images in $D$ |
| $x_i$: $i\text{-}th$ image of $D$ |
| $n$: number of pixels of $x_i$ |
| $x_i^j$: $j\text{-}th$ feature of the $i\text{-}th$ image |
| $z_i$: true class of $x_i$ |
| $D^z$: dataset containing the images of $D$ assigned by $\Phi$ to class $z$ |
| $\varphi^z$: Deep SAD model for class $z$ |
| $c_z$: center of the embedding space defined by $\varphi^z$ |
| $d$: distance |
| $Th_z$: threshold of the distance $d$ used to distinguish anomalous from normal explanations of class $z$ |
| **Other Symbols:** |
| $h$: feature maps of a selected convolutional layer |
| $U \times V$: spatial dimension of $h$ |
| $\alpha_k^z$: importance of the $k-th$ feature map with respect to class $z$ |
| $s^*$: mask that highlights the regions of the input image that have a significant positive influence on the output |
| $D_{UL}$: Dataset of unlabeled samples |
| $D_{LA}$: Dataset of samples labeled as anomalous |
| $D_{LN}$: Dataset of samples labeled as normal |
| $T_{UL}$: number of unlabeled samples |
| $T_{LA}$: number of samples labeled as anomalous |
| $T_{LN}$: number of samples labeled as normal |
| $W$: weights of the Deep SAD model |
| $\eta$: hyperparameter used to weight the term of the Deep SAD loss function associated to labeled samples |
| $\mu$: faithfulness metric |
| $\underline{x}$: baseline value for the features of $x$ |

Significant progresses have been made in recent years in the field of eXplainable Artificial Intelligence (XAI), where several techniques have been proposed to explain DL models and understand how their outputs are obtained [8]. Specifically, in fault diagnostics, explanations can help experts identify samples where the DL model makes inferences based on irrelevant parts of the input, thereby indicating that the DL model's output should not be trusted since it may have been generated by shortcuts or biases. For this task, XAI methods that highlight the most important input features for the classification of a given sample (local explanations by feature attribution [9]) and are applicable to any DL model (model-agnostic [9]), such as SHAP [10], LIME [11], LRP [12], GradCAM [13], Counterfactuals [14], [15], CartoonX [16], have been widely used.

Research studies indicate a synergistic effect when humans and AI models collaborate, especially if humans are provided not only with the output of the AI model but also with explanations on how the algorithm obtained that output [17]. For example, a deception detection task was used as a testbed to investigate how explanations and predictions of DL models can improve human performance and foster more trustworthiness in the DL model [18]. However, in the context of fault diagnostics in large infrastructures by image processing, evaluating each explanation by domain experts is time-consuming and prone to errors [19]. This calls for the development of tools for the automatic analysis of explanations to reduce the burden on domain experts. Despite the extensive research in the field of XAI, most existing studies focus primarily on generating explanations to build user trust. In contrast, only a limited number of XAI methods, such as SprAy [20], faithfulness metric [21], Test with Concept Activation Vectors (TCAVs) [22], Network Dissection [23] and Automated Concept-based Explanations (ACE) [24], which will be discussed in Section 2.1, have been developed for the automatic analysis of these explanations [25]. These methods face significant challenges: SprAy is proposed for explaining samples of the training set, which prevents its use for supporting maintenance, whereas concept-based methods (TCAVs, Network Dissection and ACE) require the involvement of field experts to define or identify concepts corresponding to abnormal behaviors of the model.

In this work, we address a common scenario in industrial applications where a DL-based model is employed for fault diagnostics in large infrastructures. The model processes images of infrastructure components as input and classifies their health states (e.g. healthy, degraded, failed). Our objective is to develop a model-agnostic framework that automatically identifies misclassified images or those classified using non-causal shortcuts by automating the analysis of their explanations. The proposed framework includes an XAI model for obtaining post-hoc explanations and then a semi-supervised AD model that processes these explanations to identify abnormal behaviors of the model. In particular, we use GradCAM [13] for generating explanations and Deep Semi-Supervised Anomaly Detection (Deep SAD) [26] for processing explanations. GradCAM is chosen among local post-hoc XAI methods due to its ability to produce explanations that highlight spatially coherent and meaningful regions, facilitating automatic processing [13]. Subsequently, Deep SAD [27] is used to identify anomalous explanations that may indicate model abnormal behaviors. This allows maintenance decision-makers to focus their efforts on manually reviewing only images flagged by Deep SAD. Deep SAD is selected for the following reasons:

- As a deep method, it has been demonstrated to outperform shallow anomaly detection approaches when handling image data [27];
- as a semi-supervised method, it can leverage available information about anomalies, such as explanations of misclassified images;
- it does not impose constraints on the compactness of clusters formed by anomalous explanations, which is essential for accommodating rare or previously unseen types of explanations.

We evaluate our framework using a case study on diagnosing faults in power grid insulators. This application is particularly significant because insulators are critical components, involved in 81.30% of power grid accidents [28], and their failure can lead to power supply interruptions [29]. Traditional manual inspection of insulators is labour-intensive, dangerous and costly [28]. Specifically, we use drone-collected images of insulators shells to classify their health states. The performance of Deep SAD in identifying misclassifications based on GradCAM explanations is compared to that of the faithfulness metric, a state-of-the-art approach for assessing explanation quality[21].

The methodological contribution of this work is to propose a novel framework, based on the combination of existing techniques, that identifies abnormal behaviours of a DL model by automatically processing explanations of test images. Unlike current state-of-the-art approaches for explanation analysis, our framework does not require intervention from model developers and field experts to identify and classify concepts. Being applicable to newly collected images, this framework practically enhances the monitoring of large infrastructures by improving the performance of a DL classifier while limiting manual reclassification by maintenance operators to only a small subset of test images for which model abnormal behaviors have been identified.

The remainder of the paper is organized as follows. In Section 2, we discuss the related works about automatic analysis of explanations and anomaly detection. In Section 3, we formulate the problem. In Section 4, we provide the necessary background on the methods used in the proposed framework. In Section 5, we outline the proposed framework. Section 6 introduces the case study and discusses the hyperparameters setting and Section 7 presents the obtained results. Section 8 draws the conclusions and outlooks on the possible future developments of the research.

## 2. Related works

### 2.1 Automatic analysis of explanations

The automatic analysis of explanations is a rapidly evolving field in XAI, driven by the need of enhancing the interpretability and reliability of DL models. Whereas many XAI methods focus on generating explanations to build user trust, only few have been developed to systematically analyze the explanations for detecting and, eventually, rectifying model abnormal behaviors. In this Section, we review several prominent post-hoc, model agnostic XAI techniques that can be used to identify abnormal behaviors in models used to support maintenance decisions.

Spectral Relevance Analysis (SprAy) is designed to automatically process explanations by identifying clusters of similar explanations, thereby discerning the different strategies adopted by the classification model [20]. SprAy utilizes an unsupervised spectral clustering algorithm and has been demonstrated to effectively identify Clever-Hans behaviors, where models rely on spurious correlations rather than meaningful features. Following the detection of such abnormal behaviors, Augmentative Class Artifact Compensation (A-ClArC) and Projective Class Artifact Compensation (P-ClArC) can be used to correct these behaviours [30]. However, SprAy is limited to processing explanations from the training set, making it unsuitable for real-time maintenance support on test data. Additionally, while A-ClArC and P-ClArC are effective at correcting Clever-Hans behaviours, they may not work to address other types of unwanted behaviors [7].

The faithfulness metric has been introduced to evaluate whether a XAI method, providing explanations in the form of feature attributions, effectively identifies important input features [21]. This involves masking parts of the test image, i.e., setting random subsets of features to a baseline value, and measuring the resulting changes in the model output; the faithfulness of an explanation is, then, computed as the correlation between the changes in output when features are masked and the importances assigned to those features by the XAI method: low faithfulness (small correlation) indicates a model abnormal behavior in the classification of the image.

Alternatively, some XAI methods leverage human-interpretable concepts to automatically identify model abnormal behaviors. For example, Network Dissection [23] associates human-interpretable concepts extracted from images with the activation states of neurons of convolutional DL models. By analyzing these activation states for test images, Network Dissection provides insights into how different concepts contribute to the model's predictions. If experts can define which concepts indicate an abnormal behaviour of the DL model for a specific task, such abnormal behaviours can be automatically identified by monitoring the corresponding neurons activation states. Testing with Concept Activation Vectors (TCAV) offers an alternative approach to identify the contribution of concepts to model decisions. It is based on the use of Concept Activation Vectors (CAVs) that represent directions in the neuron activation space corresponding to specific concepts [22]. In TCAV, experts define a concept and label a set of images as containing or not such concept; then, a linear classifier is trained to distinguish between neuron activation states of samples that contain the concept and those that do not. By computing the derivative of the model output with respect to the CAV, TCAV assesses the extent to which the model relies on the concept for its decisions. During test, abnormal behaviors are detected when the activations of concepts corresponding to abnormal behaviours are large. Notable limitations of TCAVs are its reliance on domain experts to define (possibly all) concepts which can correspond to abnormal behaviors for a specific task, and the cost of labelling images for training CAVs. Automatic Concept-Based Explanations (ACE) partially addresses these issues by automatically extracting concepts from images [24]. ACE identifies representative concepts by segmenting images at different resolution levels and clustering the obtained segments. Each concept is, then, associated with a CAV [22] and, as done for TCAVs, used to detect potential model abnormal behaviors. Although ACE automates the extraction of concepts, expert knowledge is still necessary to determine which concepts indicate a model abnormal behavior for the desired task. Also, if the set of extracted concepts is not complete, new abnormal behaviors based on missing concepts, cannot be detected during test.

Despite these advancements, existing methods present significant limitations, particularly regarding their applicability to test data for real-time maintenance support and their dependence on expert intervention for identifying and classifying concepts. These limitations highlight the need for more robust and automated approaches to the analysis of XAI explanations, which our proposed framework aims to address.

Table 2 summarizes state-of-the-art methods, highlighting their applicability for online maintenance decision-making, the type of input data that they process, the output that they provide, and the efforts required by field experts and model developers for their development.

Table 2: Summary of the methods for the automatic analysis of explanations to identify abnormal behaviors of DL models

| Method | Applicable to test images | Effort required to domain experts | Input | Output |
|---|---|---|---|---|
| SprAy [20] | NO | Identify clusters of explanations indicating a model abnormal behavior | Explanations of the images of the training set | Clusters of similar explanations |
| Network Dissection [23] | YES | Determine whether the contribution of identified concepts indicates a model abnormal behavior | Additional dataset of images corresponding to concepts | Correspondence between internal activations and concepts |
| TCAVs [22] | YES | Define concepts and label images as containing or not such concepts | Set of images labelled as containing or not containing the target concept | Sensitivity of the classifier output to the target concept |
| ACE [24] | YES | Determine whether the contribution of identified concepts indicates a model abnormal behavior | Set of images labelled with their true class | Concepts used to classify the images of a given class they can be used as input to TCAVs |
| Faithfulness [21] | YES | No effort | Image and corresponding explanation | Faithfulness of the explanation |

## 2.2 Anomaly detection

Anomaly Detection (AD) identifies unusual or rare samples within a dataset [26]. The most common AD approaches are unsupervised, operating under the assumption that the majority of the available samples represent normal behavior, whereas anomalous samples are infrequent and deviate from this norm. Among the widely recognized unsupervised AD methods are Isolation Forest (IF) [31], Local Outlier Factor (LOF) [32] and Copula-based Outlier Detection (COPOD) [33]. IF is a tree-based method that detects anomalies by recursively partitioning the data. It randomly selects a feature and a split value within the range of that feature until each sample is isolated or a predefined maximum tree depth is reached. The number of splits required to isolate a data point is, then, used to compute its anomaly score. Anomalies are expected to be isolated with fewer splits compared to normal data due to their distinct attributes. Local Outlier Factor (LOF) is a density-based AD method that identifies outliers by comparing the local density of a sample to the local densities of its neighbours. It relies on computing distances between each point and its $k$ nearest neighbours to measure the Local Reachability Density (LRD). The LOF score for each point is calculated as the ratio of the average LRD of its $k$ nearest neighbours to the LRD of the point itself. A point is identified as an anomaly if its LOF score is significantly greater than 1, indicating that its local density is substantially lower than that of its neighbours. COPOD is a non-parametric method that employs copula functions to model the joint distribution of data features, detecting anomalies as observations with low estimated probabilities. COPOD effectively captures complex dependencies between features without making restrictive assumptions about the underlying data distribution. Alternatively, a common approach is to learn a "compact" representation of normal data. One-Class Classification (OCC) estimates the distribution of normal data by defining a binary function to segregate normal from anomalous samples [34]. Instead of seeking the entire distribution of normal data, a simpler approach is to identify the boundary that separates normal from anomalous samples [35]. When the decision boundary is complex in the original space, the problem can be simplified by mapping the data in a larger dimensional feature space, where the decision boundary becomes simpler, by using kernels [36]. For example, One-Class Support Vector Machine (OCSVM) employs a hyperplane as the decision boundary to separate normal samples from the origin by maximizing their distance from the origin [37]. Alternatively, the Support Vector Data Description (SVDD) method encloses normal data within the smallest possible hypersphere[38].

The shallow approaches described above, often require manual feature engineering to effectively handle high-dimensional data and are difficult to scale for large datasets. These challenges have sparked great interest in developing novel, deep approaches to AD, such as Deep Support Vector Data Description (Deep SVDD) [39], which integrates deep neural networks to capture compact data representations along with the SVDD principle to define the boundary between normal and anomalous samples. Furthermore, when few anomalies are available in the training set semi-supervised learning can enhance AD performance. In this work we resort to Deep SAD [26], which is a deep semi-supervised anomaly detection method that combines the representation capabilities of deep approaches with the information of both normal and anomalous samples available in the training set.

## 3 Problem formulation

Monitoring large infrastructures with DL models poses to maintenance decision-makers the challenge of validating the classification of a large number of images to make informed decisions. Whereas XAI methods can explain the output of DL models, they typically produce one explanation per image, making it impractical for maintenance decision-makers to review each explanation individually.

In this context, the objective of this work is to develop a framework for processing explanations to support maintenance decisions. Specifically, we aim to automatically assess whether the classification of the DL model should be revised before a maintenance decision, based on the explanation provided by the XAI method.

The problem is formulated assuming the availability of a set $D = \{x_i, z_i\}_{i=1,..,T}$ of $T$ images $x_i$ of components of a large infrastructure, and the corresponding classes $z_i \in \{1,..,Z\}$, corresponding to the components health state (e.g. healthy, degraded, failed). Each image is described by a vector of $n$ features $[x_i^1, ..., x_i^n]^T$.

For each test image $x$, we predict its class $\hat{z}$ using a DL classification model $\Phi$: $\hat{z} = \Phi(x): \mathcal{R}^n \to \{1,..,Z\}$. Then, a corresponding explanation $y$ is obtained from a XAI model $g$: $y = g(\Phi, x)$ and the aim is to automatically assess whether the explanation $y$ is normal ($\hat{l}(y) = 1$), i.e. it is similar to explanations of other images of the same class $z$ correctly classified by $\Phi$, or anomalous ($\hat{l}(y) = -1$): in the latter case, the image and its explanation are reviewed for potential reclassification before maintenance decision.

# 4 Background

## 4.1 Pre-trained classification models

In this Section we briefly describe the CNN architectures utilized in this work to validate the proposed framework: MobileNetV3 Small [40] and EfficientNet-B0 [41]. These networks are chosen for their computational efficiency and high accuracy in image classification tasks, making them well-suited for large-scale infrastructure monitoring applications.

MobileNetV3 Small is a lightweight CNN architecture [40] that builds upon previous MobileNet versions [42] by incorporating advancements from neural architecture search (NAS), an automated technique for designing neural networks architectures that optimize performance metrics such as accuracy and computational efficiency [43]. A key feature of MobileNetV3 Small is the use of depthwise separable convolutions, which decompose standard convolutions into a depthwise and a pointwise convolution. This decomposition significantly reduces the number of parameters and the computational cost without compromising accuracy. Additionally, MobileNetV3 Small integrates squeeze-and-excitation (SE) blocks, improving feature recalibration by dynamically weighting channel-wise feature maps [44]. The architecture also employs the hard-swish (H-swish) activation function, a computationally efficient approximation of the swish activation function defined as $swish(x) = x \cdot sigmoid(x)$ [45]. The H-swish activation maintains the performance benefits of swish while reducing the computational complexity associated with the sigmoid function, which can be inefficient to compute [46].

EfficientNet-B0 serves as the baseline model of the EfficientNet family, which is renowned for its novel scaling method that uniformly scales the network's width, depth and resolution using a fixed compound coefficient [41]. This balanced scaling approach results in a more compact and efficient model that maintains, or even improves, accuracy compared to traditional scaling methods. EfficientNet-B0 uses mobile inverted bottleneck convolutions, a building block that extends depthwise separable convolutions by incorporating SE blocks. This enhancement improves the network's representational capability without adding significant computational overhead.

Both MobileNetV3 Small and EfficientNet-B0 have been extensively validated on several benchmark datasets, demonstrating their capability to achieve efficient inference while maintaining high classification accuracy. Their proven performance and resource efficiency make them well-suited for deployment in real-time maintenance and monitoring systems for large infrastructures, ensuring reliable and fast fault diagnostics.

## 4.2 GradCAM

GradCAM [13] is a local, model-specific XAI method that provides visual explanations for image classifications by leveraging the internal representations of CNNs. It estimates the importance of the image pixels in contributing to the model's output classification by propagating gradients through the network.

Consider a classification model $\Phi$ that processes an input $x$ and assigns it the score $\Phi_z(x)$ (before the Softmax activation) for a target class $z$. Let $h = [h^1, \ldots, h^k, \ldots, h^K]$ represent the feature maps from a selected convolutional layer, typically the last one due to its ability to produce coarse explanations [13]. The spatial dimensions of each feature map $h^k$ are $U \times V$, which are generally smaller than the dimensions of the input image due to the nature of convolutional operations. The importance $\alpha_k^z$ of the $k-th$ feature map with respect to class $z$ is calculated by averaging the gradients $\frac{\partial \Phi_z(x)}{\partial h^k}$ of the class score with respect to the feature map over the spatial dimensions:

$$\alpha_k^z = \frac{1}{U \times V} \sum_{u=1}^{U} \sum_{v=1}^{V} \frac{\partial \Phi_z(x)}{\partial h_{u,v}^k}$$

These weights $\alpha_k^z$ quantify the contribution of each feature map $h^k$ to the target class $z$. GradCAM then constructs a coarse localization map $s^* \in R^{U \times V}$ by performing a weighted combination of the feature maps, followed by applying the Rectified Linear Unit (ReLU) activation function to retain only the positive influences:

$$s^* = ReLU\left(\sum_{k=1}^{K} \alpha_k^z h^k\right)$$

The ReLU activation function ensures that only regions with a positive contribution to the score are considered. The resulting mask $s^*$ highlights the regions of the input image that have a significant positive influence on the model's prediction for class z. Since the dimensions of $s^*$ are $U \times V$, a resizing step is necessary to match the original input image dimensions. This is typically done through interpolation, allowing the importance scores to be mapped back to the input pixels of image $x$. The final GradCAM heatmap effectively visualizes the areas of the image that the model considers most relevant for its classification decision. The implementation of GradCAM is summarized in Algorithm 1 below.

---

**Algorithm 1** GradCAM to compute explanation $s^*$

**Input:** Image $x \in \mathbb{R}^n$, trained classifier $\Phi$, target class z

**Forward Pass:**
Compute the model output $\Phi(x)$
Compute the score (before Softmax) of the target class: $\Phi_z(x)$
Extract feature maps $h = [h^1, h^2, \ldots, h^K]$ from a selected convolutional layer

**Backward Pass:**
Backpropagate gradient to compute $\frac{\partial \Phi_z(x)}{\partial h^k}$ for all $k$

**Compute Importances:**
For each feature map $h^k$, compute importance $\alpha_k^z$:

$$\alpha_k^z = \frac{1}{U \times V} \sum_{u=1}^{U} \sum_{v=1}^{V} \left( \frac{\partial \Phi_z(x)}{\partial h_{u,v}^k} \right)$$

**Build Mask:**
Compute $s^* = \text{ReLU}\left( \sum_{k=1}^{K} \alpha_k^z h^k \right)$

**Reshape Mask:**
Resize $s^*$ to match the dimensions of the input image $x$

**Output:** Mask $s^* \in \mathbb{R}^n$ highlighting important regions of $x$ for target class $z$

---

Algorithm 1: GradCAM

### 4.3 Deep SAD

Deep SAD is a semi-supervised anomaly detection method that allows exploiting the information of both unlabeled and labeled samples [26], to improve AD performance with respect to fully unsupervised methods, when only a limited set of anomalies is available for training [47]. Deep SAD employs a DL model to map input samples into an embedding space, with the objective of minimizing the distance $d$ between the center of this space and normal or unlabeled samples, while maximizing $d$ between the center and anomalous samples. In accordance with the problem formulation of this work, we introduce Deep SAD with respect to the specific task of identifying anomalous samples among explanations. Specifically, we consider a set of $T_{UL}$ unlabeled samples $D_{UL} = \{y_i\}_{i=1,\ldots,T_{UL}}$, a set of $T_{LN}$ samples labeled as normal $D_{LN} = \{y_i\}_{i=1,\ldots,T_{LN}}$ and a set of $T_{LA}$ samples labeled as anomalous $D_{LA} = \{y_i\}_{i=1,\ldots,T_{LA}}$. The weights $W$ of the deep neural network $\varphi$ defining the embedding space are set by minimizing the loss function:

$$\mathcal{L}_{Deep\ SAD}(W) = \mathcal{L}_{UL} + \eta \times (\mathcal{L}_{LA} + \mathcal{L}_{LN}) \quad (1)$$

with:

$$\mathcal{L}_{UL} = \frac{1}{T_{UL} + T_{LN} + T_{LA}} \left( \sum_{y_i \in D_{UL}} \| \varphi(y_i; W) - c \|^2 \right)$$

$$\mathcal{L}_{LA} = \frac{1}{T_{UL} + T_{LN} + T_{LA}} \left( \sum_{y_i \in \mathcal{D}_{LA}} (\| \varphi(y_i; W) - c \|^2)^{-1} \right)$$

$$\mathcal{L}_{LN} = \frac{1}{T_{UL} + T_{LN} + T_{LA}} \left( \sum_{y_i \in \mathcal{D}_{LN}} \| \varphi(y_i; W) - c \|^2 \right)$$

where $c$ is the center of the embedding space and the hyperparameter $\eta > 0$ weights the importance of the term related to the labeled samples. The $\mathcal{L}_{UL}$ term focuses on unlabeled samples, aiming to minimize their distance $d$ from the center $c$, whereas the $\mathcal{L}_{LA}$ term considers samples labeled as anomalous, seeking to maximize the distance $d$ of anomalous samples from $c$ and $\mathcal{L}_{LN}$ term considers samples labeled as normal whose distance should be minimized. The main steps of Deep SAD models are summarized in Algorithm 2 below.

**Algorithm 2** Deep SAD to compute anomaly scores

**Data:**
Unlabeled samples $\mathcal{D}_{UL} = \{y_i\}_{i=1}^{T_{UL}}$
Labeled normal samples $\mathcal{D}_{LN} = \{y_i\}_{i=1}^{T_{LN}}$
Labeled anomalous samples $\mathcal{D}_{LA} = \{y_i\}_{i=1}^{T_{LA}}$

**Parameters:**
Deep neural network $\varphi$ with weights $W$
Hyperparameter $\eta > 0$

**Initialization:**
Initialize weights $W$
Set center $c$ (forward pass of normal samples)

**Training:**
for each epoch do
    for each batch $\mathcal{B}$ of samples do
        Initialize loss $L = 0$
        $N \leftarrow |\mathcal{B}|$     ▷ $N$ is the total number of samples in the batch
        for each $y_i \in \mathcal{B}$ do
            Compute Loss:
            $L = L + \|\varphi(y_i; W) - c\|^2 + \eta \times \left[ \|\varphi(y_i; W) - c\|^2 + (\|\varphi(y_i; W) - c\|^2) \right]^{-}$
        end for
        $L \leftarrow L/N$
        Update weights $W$ by minimizing $L$ using Adam optimizer
    end for
end for

**Output:** Trained network $\varphi(y; W)$ and center $c$ to compute anomaly scores

Algorithm 2: Deep SAD

## 5. Proposed framework

Figure 1 illustrates the proposed framework, which is based on a supervised DL classifier $\Phi$, an explanation model $g$ and an anomaly detection model $\Psi$.

The classifier $\Phi$ determines the health state $\hat{z}$ of the infrastructure component depicted in the image (e.g., healthy, degraded, failed). Two convolutional neural network (CNN) architectures are considered in this work for image classification: MobileNetV3 Small [40] and EfficientNet-B0 [41] (Section 4.1).

Concerning the explanation model $g$, GradCAM is chosen because it generates explanations $y = g(\Phi, x)$ in the form of relevance maps that highlight the pixels most influential in the class assignment, providing an interpretable visual representation of the features that most contributed to the classification (Section 4.2). The label $\hat{z}$ assigned by $\Phi$ is used as target class for GradCAM computation, ensuring that the explanations are coherent and consistent with model $\Phi$'s classifications.

The AD model $\Psi$ processes explanations to classify them as either normal ($\hat{l} = 1$) or anomalous ($\hat{l} = -1$). In the latter case, the image and its explanation are reviewed by maintenance operators for potential reclassification or identification of non-causal shortcuts.

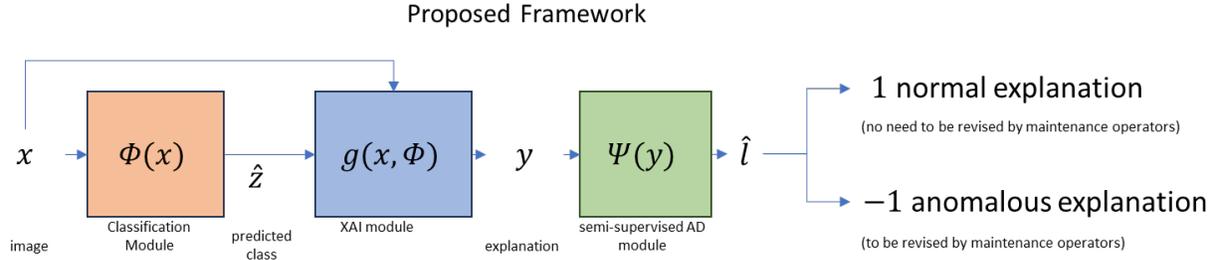

Figure 1: proposed framework

Figure 2 shows the semi-supervised AD module used in this work, which consists of $Z$ Deep SAD models (Section 4.3), denoted as $\varphi^z$ for each class $z = 1, \ldots, Z$. Each Deep SAD model $\varphi^z$ focuses on explanations of images assigned to class $z$ by $\Phi$. The Deep SAD models operate by mapping explanations into an embedding space specific to each class. In this space, explanations of correctly classified images form a compact cluster around a central point $c_z$, whereas explanations of incorrectly classified images are dispersed away from these clusters. For a given explanation $y$, the Deep SAD model $\varphi^z$ computes the distance $d(c_z, y)$ between the embedding of $y$ and the center $c_z$ of the embedding space of the embedding space for class z. This distance $d(c_z, y)$ is then compared to a predefined threshold $Th_z$ to determine whether the explanation is normal ($\hat{l} = 1$) or anomalous ($\hat{l} = -1$):

$$\Psi(y) = \begin{cases} \hat{l} = 1 \text{ if } d(y, c_z) < Th_z \\ \hat{l} = -1 \text{ if } d(y, c_z) \geq Th_z \end{cases}$$

Images corresponding to anomalous explanations are reviewed by maintenance operators to rectify possible misclassifications and identify non-causal shortcuts. The decision to develop $Z$ distinct models $\varphi^z$ for identifying anomalous explanations rather than a single model is driven by the observation that explanations of images correctly classified into different classes often form distinct clusters. This diversity can make detecting anomalous explanations using a single AD model more challenging.

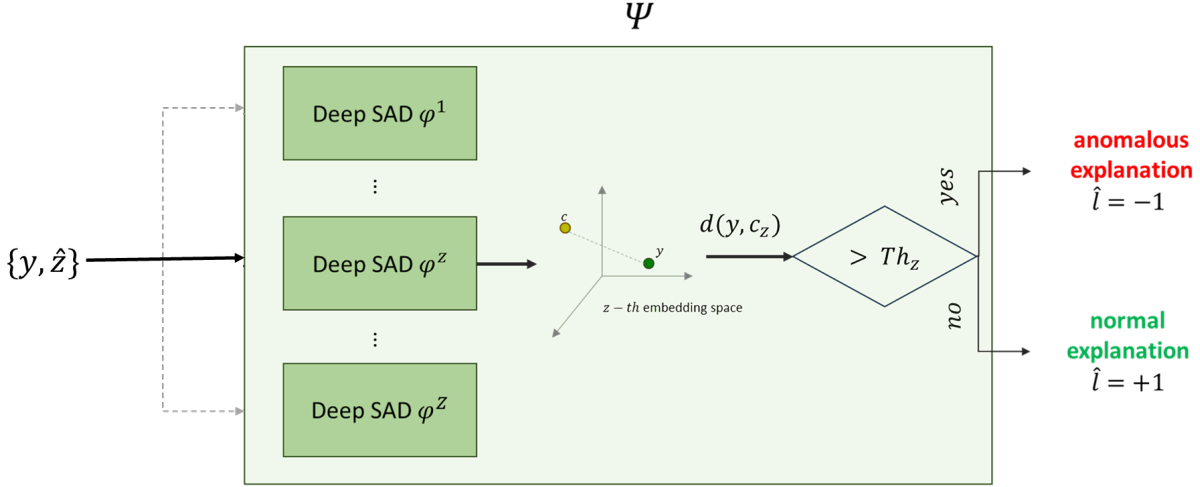

Figure 2: the semi-supervised AD module receives in input the explanation $y$ and the predicted class $\hat{z}$. It consists of $Z$ Deep SAD models $\varphi^z$ that map explanations in an embedding space where the distance is compared with a threshold to determine if they are normal or anomalous.

The model $\Psi$ is developed following the steps reported in Algorithm 3. It uses the dataset $D = \{(x_i, z_i)\}_{i=1,\ldots,T}$ formed by $T$ images $x_i$ and the corresponding class $z_i$, then, the following steps are performed:

1) Classify the $T$ images $x_i$, $i = 1, \ldots, T$, with $\Phi$, i.e. $\hat{z}_i = \Phi(x_i)$, dividing them into $Z$ datasets $D^z \subseteq D$, each formed by images $x_i$ assigned to class $z$: $D^z = \{x_i \in D \mid \hat{z}_i = z\}$.
2) Compute the explanation $y_i = g(\Phi, x_i)$ of each image $x_i \in D^z$ using GradCAM.
3) Divide the dataset $D^z$ is into a training set $D^z_{train}$ and a validation set $D^z_{val}$. $D^z_{train}$ is used to develop the class-specific Deep SAD model $\varphi^z$. The computation of the Deep SAD loss function (Eq. 1) requires defining the three sets $D_{UL}$, containing explanations that we do not know if they are normal or abnormal, $D_{LA}$ containing anomalous explanations and $D_{LN}$ containing normal explanations. Specifically, in this work we assume that explanations of the correctly classified images are unlabelled because some images might be correctly classified using non-causal shortcuts, and, therefore, they correspond to anomalous explanations:

$$D_{UL} = \{y_i \in D^z_{train} \mid \hat{z}_i = z_i\}$$

explanations of images incorrectly classified by $\Phi$ are anomalous:

$$D_{LA} = \{y_i \in D^z_{train} \mid \hat{z}_i \neq z_i\}$$

whereas $D_{LN} = \emptyset$, since explanations are not a priori identified as normal or anomalous. The expectation is that a small fraction of the images is incorrectly classified due to the typically high accuracy of DL classifiers, so that the typical AD setup is satisfied. Deep SAD's training does not focus on the compactness of anomalous explanations, recognizing the diverse nature of misclassifications and shortcuts. Following the approach recommended in [39], Deep SAD is pre-trained by an autoencoder whose objective is to reconstruct the explanations $y_i$, precisely as they are. This process yields a data representation with reduced dimensionality at the encoder's bottleneck. Subsequently, the encoder is fine-tuned using the loss function of Eq. 1. The center $c_z$ of the embedding space is initialized by computing the average from a forward pass over the training data $D^z_{train}$.

To set the threshold $Th_z$, the trained Deep SAD model $\varphi_z$ is applied to the explanations $y_i$ of the validation set images $x_i \in D^z_{val}$, and the distances $d(y_i, c_z)$ of the explanations $y_i$ from the embedding space center $c_z$ are computed. The optimal threshold $Th_z$ maximizes the $F_\beta$ score of the validation set, considering anomalous explanations ($\hat{l}_i = -1$) as misclassifications and normal ones ($\hat{l}_i = 1$) as correct classifications by $\Phi$. The $F_\beta$ score [37] provides a trade-off between precision and recall in imbalanced binary classification tasks [48] using the parameter $\beta$ to weight the importance of the two terms.

$$F_\beta = (1+\beta^2) \frac{precision \times recall}{(\beta^2 \times precision) + recall} \tag{2}$$

Precision is calculated as the ratio of explanations correctly identified as anomalous ($\hat{l}_i = -1 \text{ and } \hat{z}_i \neq z_i$) to the total number of explanations identified as anomalous ($\hat{l}_i = -1$), while recall is the ratio of explanations correctly identified as anomalous ($\hat{l}_i = -1 \text{ and } \hat{z}_i \neq z_i$) to the actual number of misclassifications ($\hat{z}_i \neq z_i$):

$$Precision = \frac{\sum_{x_i \in D^z_{val}} I(\hat{l}_i = -1 \cap \hat{z}_i \neq z_i)}{\sum_{x_i \in D^z} I(\hat{l}_i = -1)}$$

$$Recall = \frac{\sum_{x_i \in D^z_{val}} I(\hat{l}_i = -1 \cap \hat{z}_i \neq z_i)}{\sum_{x_i \in D^z} I(\hat{z}_i \neq z_i)}$$

where $I$ is the indicator function, which is equal to 1 if the condition is true, and to 0 otherwise.

**Algorithm 3** Proposed Framework for Identifying Anomalous Explanations

**Input:**
Dataset $D = \{(x_i, z_i)\}_{i=1}^T$; $x_i$ are images and $z_i$ are labels
Trained classifier $\Phi$, explanation model $g$
**Methodology Development:**
    ▷ 1. Classify images
**for** each image $x_i$ in $D$ **do**
    Compute predicted label $\hat{z}_i = \Phi(x_i)$
**end for**
    ▷ 2. Group images by predicted class:
**for** each class $z$ in $\{1, 2, \ldots, Z\}$ **do**
    $D^z \leftarrow \{x_i \in D \mid \hat{z}_i = z\}$
**end for**

**for** each class $z$ in $\{1, 2, \ldots, Z\}$ **do**
    **for** each image $x_i$ in $D^z$ **do**
        ▷ 3. Get GradCAM explanations (Algorithm 1):
    $y_i = g(\Phi, x_i)$
    **end for**

    **for** each fold of the cross-validation **do**
        Split $D^z$ into $D^z_{train}$ and $D^z_{val}$
        ▷ 4. Train Deep SAD model (Algorithm 2):
        $D_{UL} \leftarrow \{y_i \in D^z_{train} \mid \hat{z}_i = z_i\}$ for unlabeled data (correctly classified)
        $D_{LA} \leftarrow \{y_i \in D^z_{train} \mid \hat{z}_i \neq z_i\}$ for anomalous data (misclassified)

        Train Deep SAD model $\varphi_z$ on $D^z_{train}$
        ▷ 5. Determine Threshold $Th_z$:
        **for** each explanation $y_i$ in $D^z_{val}$ **do**
            Map explanation in embedding space $\varphi_z(y_i)$
            Compute distance from the center $d(\varphi_z(y_i), c_z)$
        **end for**
        Find $Th_z$ that maximizes $F_\beta$ score on $D^z_{val}$ (refer to Equation (2))
    **end for**
**end for**
**Output:** Set of $Z$ models for identifying anomalous explanations

Algorithm 3: proposed framework

## 6. Case study

This work explores fault diagnostics in power grids using images collected by drones. Drones offer novel infrastructure monitoring solutions, covering extensive areas quickly and capturing high-resolution images or videos, thus enabling more efficient inspections compared to traditional methods. Additionally, drone inspections enhance safety and reduce costs by minimizing the need for personnel in hazardous or hard-to-reach areas [49]. However, the large amount of images that can be collected in short intervals of time poses challenges to the automation of their analysis. Specifically, this study focuses on diagnosing faults in insulators, which are components crucial for maintaining insulation resistance and preventing power supply interruptions [29].

We consider the openly available dataset Insulator Defect Image Dataset (IDID), that can be downloaded at [50]. The dataset contains high-resolution images of power line insulators (Figure 3a), with the bounding boxes identifying the insulator shells and a categorization of the health state of the individual shells into the three health states ($Z = 3$): broken, flashover and healthy [50]. Sub-images of individual shells are extracted from each insulator image using the bounding boxes and categorized based on provided labels. The dataset displays an imbalance across the three health states, with images of 13,336 healthy, 2,564 flashover, and 1,204 broken shells. To accommodate varying dimensions and shapes, images are adjusted to 256x256 pixels using black padding. An example from each class is illustrated in Figure 3 (b, c and d).

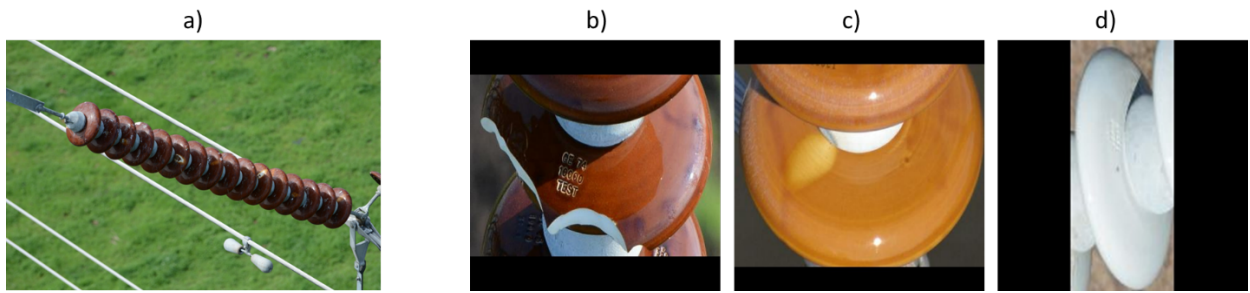

Figure 3a): An image showcasing power line insulators. b), c) and d): Representative images of individual insulator shells categorized as broken, flashover and healthy, respectively.

### 6.1 Model development and hyperparameters configuration

The available dataset is partitioned into two distinct datasets $D^\Phi$ and $D^\Psi$ to carry out the training and evaluation processes, respectively. Specifically, $D^\Phi$ comprises 70% of the images and is used for training the classification model $\Phi$, whereas the remaining 30% of images constitute the dataset $D^\Psi$ for developing the Deep SAD models and assessing the overall performance of the proposed framework. To ensure a fair and unbiased evaluation, all images from a same insulator are kept within the same dataset, thereby preventing data leakage and ensuring that the models are tested on fully unseen instances. To demonstrate the generalizability of the proposed framework across different DL architectures, two CNN models are employed for the classification module: MobileNetV3 Small [40] and EfficientNet-B0 [41], denoted $\Phi^M$ and $\Phi^E$ respectively. Both models are initialized with weights pre-trained on the ImageNet dataset [51], leveraging transfer learning to enhance their performance on the specific fault diagnostic task. The final fully connected layer of each model is modified to accommodate a three-class classification scenario, corresponding to the health states of the insulators shells (healthy, broken, flashover).

For GradCAM hyperparameter configuration, we follow the recommendations outlined in [13]. We select the last convolutional layer of the CNN as the representation layer to balance high-level semantic information with sufficient spatial resolution, facilitating accurate localization of important pixels in the input image. The class used for generating image explanations corresponds to the class predicted by the model.

We evaluate four Deep SAD models $\varphi^z$, each corresponding to a specific fault class ($z = $ broken or $z = $ flashover) and classifier ($\Phi^M$ or $\Phi^E$). Performance evaluation for each $\varphi^z$ employs a 5-fold cross-validation procedure, utilizing only the images from $D^\Phi$ and $D^\Psi$ assigned to class $z$. These subsets are referred to as $D^{\Phi z}$ and $D^{\Psi z}$, respectively. In each fold, $\varphi^z$ is trained on all images from $D^{\Phi_z}$ and 60% of the images of $D^{\Psi_z}$. The remaining 40% of $D^{\Psi_z}$ is split equally, with 20% of the images used to set the threshold and 20% for evaluating $\varphi^z$ performance.

For setting the hyperparameters of the Deep SAD models, we generally apply the settings reported in [26]. The four Deep SAD models share the same architecture based on the VGG16 network, which has obtained strong performances in classification tasks [52]. The weights of the VGG16 model are initialized with pre-trained ImageNet weights. The last fully connected layer of the network is modified to achieve an embedding space dimension of 10. Such a small value is chosen to mitigate overfitting, given the limited size of the training dataset compared to those used in [26]. Note that Deep SAD has been shown to be robust to variations in embedding space dimensionality [26]. Following [26], an autoencoder is used for pretraining. During testing, the threshold used for classifying an explanation as normal (close to the centre) or anomalous (far from the centre) is set by optimizing the $F_\beta$ score with $\beta$ equal to 0.1. By prioritizing precision over recall, this setting minimizes the reclassification workload for maintenance decision-makers. Table 3 provides setting of additional hyperparameters for the processing of explanations of images assigned to the flashover class by $\Phi^E$. Hyperparameters settings for the other combinations of classifiers ($\Phi^M$ or $\Phi^E$) and fault classes (broken or flashover) are reported in Appendix A, which also includes examples of loss convergence during training. Weight decay within the Adam optimizer is used for achieving regularization during minimization of the Deep SAD loss function [53]. The weight $\eta$ associated with anomalous explanations in (Eq.1) is increased from 1 (as used in [26]) to 10, due to the low number of anomalies in the Deep SAD training data, which correspond to the misclassifications by $\Phi$.

Table 3: Setting of the hyperparameters of the Deep SAD model for class flashover and classifier $\Phi^E$

| | |
|---|---|
| Number of epochs for training the autoencoder = 100 | Autoencoder Early Stopping = 20 epochs |
| Number of epochs for training the Deep SAD = 1000 | Early Stopping = 50 epochs (after epoch 100) |
| Weight decay = $10^{-5}$ | Learning Rate = $10^{-3}$ |
| Batch size = 32 | Weight applied to labeled samples in Eq. 1: $\eta = 10$ |

Training Deep SAD required 5 hours on an NVIDIA Tesla T4 GPU with 16 GB of GDDR6 memory, featuring 2,560 CUDA cores and 320 Turing Tensor Cores to accelerate deep learning computations. Evaluating a test image takes approximately 0.4 seconds on an 11th Gen Intel(R) Core(TM) i7-1165G7 @ 2.80GHz-1.69 GHz processor, which meets the application's requirements.

Our framework requires configuring a large number of hyperparameters, including those for the CNN classifier, Grad CAM, Deep SAD, and the threshold $\beta$. However, the present study aims at demonstrating the applicability of the proposed framework for supporting maintenance, rather than optimizing its performance. Further optimizations and evaluations with different models for the different modules will be considered in future research.

To streamline the computation of faithfulness, we employed the following hyperparameters: the image resolution was set to 16x16 pixels, as described in [21]. Feature subsets were defined by masks covering 20% of the 256 super-pixels (16x16), and the baseline value for masked features was set to zero. Faithfulness was computed by randomly sampling 10000 subsets for each explanation to ensure the stability of the calculated metric. After assigning a faithfulness score to each explanation, the AD task followed the same procedure as Deep SAD. Specifically, a validation set was used to set the threshold by maximizing the $F_\beta\ score\ (\beta = 0.1)$, and the faithfulness of newly collected images was compared against the threshold to determine if the explanation was normal or anomalous.

## 7. Results

Section 7.1 presents the results of the classification models $\Phi^M$ and $\Phi^E$, whereas Sections 7.2 and 7.3 present the results of the proposed framework, analysing flashover class samples in one test fold and samples of all classes of fault across all cross-validation folds, respectively.

### 7.1 Classification results

Figure 4 displays the confusion matrices obtained from the two classification models $\Phi^M$ and $\Phi^E$ on the images of $D^\Psi$.

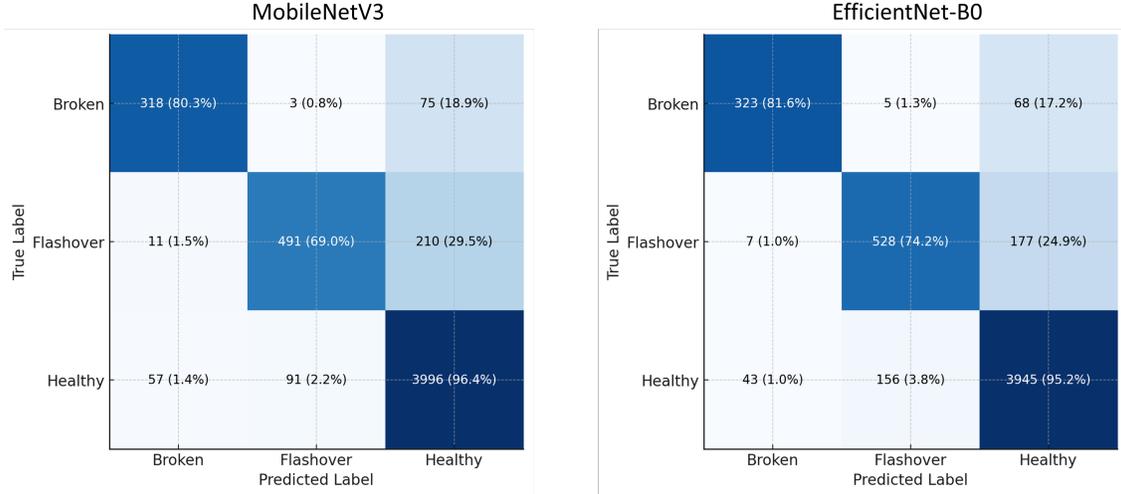

Figure 4: confusion matrix of the classifiers $\Phi^M$ (left) and $\Phi^E$ (right) on the images of $D^\Psi$. The recall (sensitivity) for each class is reported into parentheses and is represented by the colour of each cell.

Both models achieve a similar overall accuracy of approximately 91%. Specifically, $\Phi^E$ outperforms $\Phi^M$ on broken class, exhibiting higher precision (86.6% vs. 82.4%), recall (81.6% vs. 80.3%) and $F_1$ score (84.0% vs. 81.3%). Conversely, $\Phi^M$ demonstrates greater precision for the flashover class, (83.9% vs 76.6%), resulting in fewer false positive predictions, whereas $\Phi^E$ achieves higher recall (74.2% vs 69.0%), capturing more true flashover instances. The $F_1$ scores for flashover class are comparable ($\Phi^M$: 75.7%, $\Phi^E$: 75.4%), indicating a balanced trade-off between precision and recall. Both models demonstrate excellent performance for the healthy class.

The relatively lower accuracy for the broken and flashover classes highlights the need of the proposed framework for improving the model's classification accuracy. This underscores the potential for improving model performance, particularly in distinguishing challenging fault classes, thereby validating the approach.

### 7.2 Deep SAD results on one test fold

We analyze the images classified by model $\Phi^E$ as belonging to the flashover class within a single cross-validation fold. The test set contains 49 images, of which 33 are correctly identified as flashover by $\Phi^E$, while the remaining 16 are misclassified images from other classes.

Figure 5 illustrates a t-SNE [54] visualization of the 10-dimensional embedding space created by Deep SAD for projecting the explanations. Three distinct groups of explanations are observable. The group closest to the center $c_{flashover}$ (small distance group, i.e. $distance < 0.4$) contains 28 correctly classified explanations and 7 misclassified explanations of images, resulting in a 20% of misclassification within this group. In the medium distance group ($0.4 < distance < 1.1$) and large distance group ($distance > 1.1$) likelihood of misclassification increases with the distance from the center, reaching 50% and 100%, respectively. This pattern confirms Deep SAD's effectiveness in organizing the embedding space such that explanations of misclassified images are positioned farther from the center. The anomaly detection threshold is set to 1.9 by applying the same Deep SAD model to the validation set with $\beta = 0.1$, resulting in 3 true positives and zero false positives. Alternative threshold settings can be achieved by increasing $\beta$, setting $\beta = 0.5$ establishes a threshold of 1.1, leading to 4 true positives and 1 false positive, while $\beta = 1$ sets the threshold at 0.2, corresponding to 10 true positives and 10 false positives.

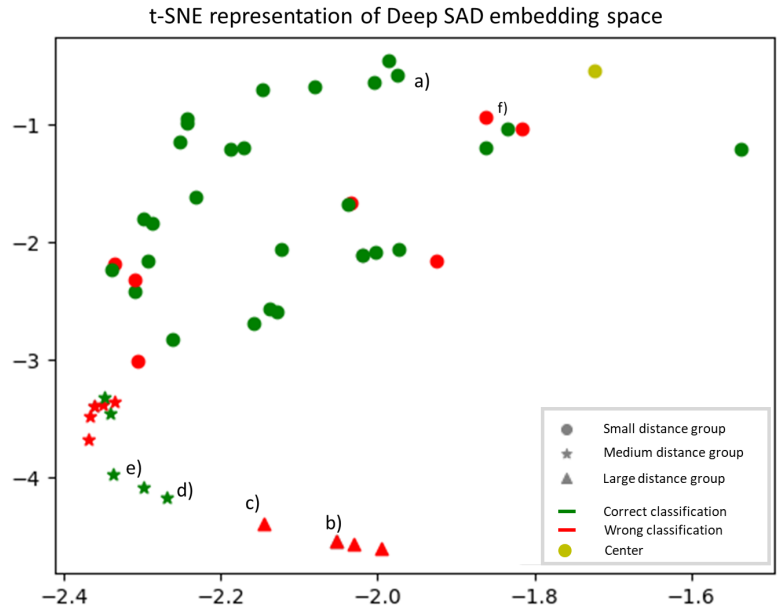

Figure 5: t-SNE representation of the embedding space. Letters a)-f) refer to the images of Figure 6

Figure 6a shows the GradCAM explanation of an image correctly classified by $\Phi^E$, which highlights the flashover-damaged region of the shell. As expected, this explanation is projected close to the embedding space's center by Deep SAD due to its similarity with other explanations of images correctly classified as flashover. In contrast, explanations 6b and 6c correspond to healthy shells that $\Phi^E$ erroneously classifies as flashover. In explanation 6b, misleading shell colours lead to the misclassification, while in explanation 6c, sunlight reflections and imprecise image framing contribute to the error. As expected, Deep SAD projects the explanations of these images, which significantly differ from those of the images correctly classified to the class flashover, far from the center of the embedding space. This separation allows maintenance operators to easily identify and review these erroneous cases. Explanations 6d, 6e and 6f require further analysis as they deviate from the typical pattern of correct and incorrect classifications based on their distance from the embedding space center. Explanation 6d illustrates a correct classification for the wrong reason. Specifically, it highlights the shell's ID tag rather than the flashover-damaged area, which is barely visible on the right side of the shell due to strong light reflections on the white surface. Consequently, Deep SAD maps this explanation far from the center, as it significantly differs from the typical flashover explanations due to the emphasis on the ID tag. This scenario presents an example of a non-causal shortcut, which underscores the importance of identifying relevant features to guide model improvements. Although revising the model's classification for this specific case does not improve overall accuracy, it offers valuable insights into potential challenges faced by $\Phi^E$. Explanation 6e involves a flashover shell correctly classified by $\Phi^E$ but projected by Deep SAD far from the centre. The occurs because GradCAM highlights nearly the entire image, failing to accurately pinpoint the flashover-damaged area. Explanation 6f shows a healthy shell erroneously assigned by $\Phi^E$ to the flashover class. In this instance, Deep SAD does not identify $\Phi'$s error, mapping the explanation close to the center. A detailed examination of the image reveals a clearer stain, which resembles flashover damage, which $\Phi^E$ uses to assign the flashover classification, as indicated by the GradCAM explanation. However, the image's low resolution (133x133) makes it challenging to determine whether these stains result from flashover damage or other reason.

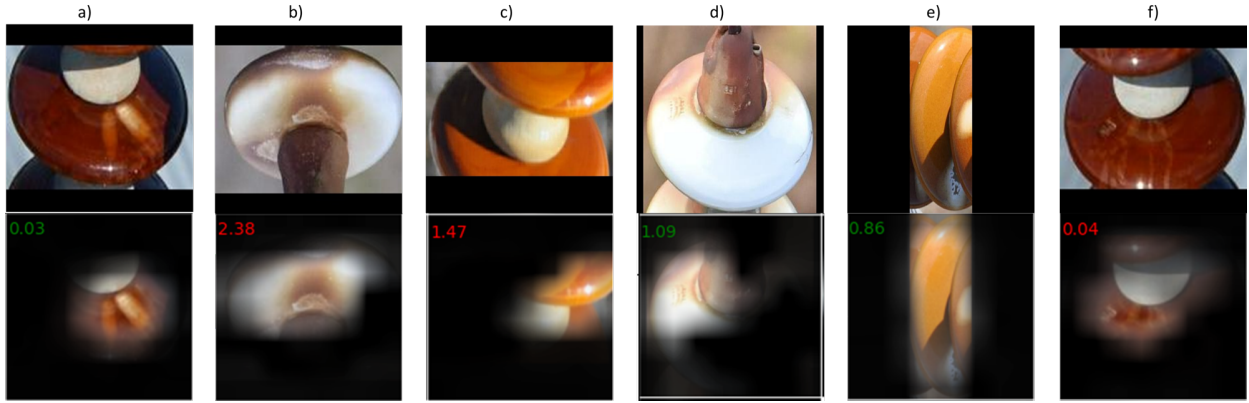

Figure 6: Images of insulators shells (top) and corresponding explanations (bottom) provided by GradCAM. In the GradCAM visualization, more important are the pixels, brighter is the color. In the upper left corner of each explanation, it is reported the distance from the center of the embedding, which is in green for correct classifications and in red for misclassifications of $\Phi^E$.

### 7.3 Results of the cross-validation

Table 4 compares the average performance of the proposed framework with the faithfulness metric across the five cross-validation folds, considering both classifiers $\Phi^E$ and $\Phi^M$. Assuming that all images flagged with anomalous explanations are correctly reclassified by the maintenance decision-makers, the proposed framework improves accuracy by 6% to 12%, while requiring maintenance operators to review 11% to 19% of the test set images. The effectiveness of the proposed framework varies across different classes, depending on the number of images incorrectly assigned by the classifier. For example, the smallest accuracy improvement (+6.35%) occurs in the classification of the broken class by $\Phi^E$, where the initial precision of $\Phi^E$ is already high (87%, Figure 4), leaving limited room for improvement. The proposed framework identifies between 24% and 50% of misclassifications, as indicated by the recall values. These relatively low recall rates are a result of selecting a $\beta$ value of 0.1, which prioritizes precision to minimize the number of images requiring maintenance operators review. Increasing $\beta$ can improve recall but may reduce precision. Precision, defined as the proportion of reviewed images that effectively need reclassification, ranges from 48% to 64%. This range includes some images correctly classified by $\Phi$ that were erroneously marked as anomalous by $\Psi$ due to shortcuts, as discussed in Section 7.2. Although reviewing these images does not improve overall accuracy, it helps identifying and addressing $\Phi'$s abnormal behaviors, informing strategies for model improvement. The overall performance of the proposed framework, measured by the $F_1$ score, remains consistent across different classes and classifiers, achieving an $F_1$ score of 0.37 in three out of four cases. The exception is the flashover class for $\Phi^M$, which achieves a higher $F_1$ score of 0.5, likely due to increased precision in this class.

The large standard deviations of the performance metrics across the five folds (e.g. standard deviations from 0.18 to 0.24 for the $F_1$ score) are primarily attributed to the small number of explanations available for the development and test of the Deep SAD model. In this scenario, minor variations in the number of anomalous explanations within the training, validation and test sets lead to significant variations of the performance metrics. Specifically, the limited number of available explanations affects the values obtained for the standard deviations of the performance metrics in two main ways: (1) it introduces instability in the Deep SAD model training process, where the choice of initial weights can significantly influence the outcome, and (2) it poses challenges in setting the anomaly detection threshold using the validation set, as the limited number of anomalous explanations may hinder generalization to the test set. Two computationally intensive solutions can be implemented to address these issues: (1) training multiple models with different random initializations of the network weights and selecting the best-performing one, and (2) employing a Leave-One-Out (LOO) cross-validation strategy to set the anomaly detection threshold with more data [55]. Implementing these strategies on $\Phi^M$ applied to the flashover class, with three different random weight initializations for each model of each fold, reduces the standard deviation of the $F_1$ score of 48% (from 0.19 to 0.10). It is worth observing that, as time passes and more images are collected, the values of the standard deviations of the obtained performance metrics are expected to decrease, even without implementing the computationally intensive solutions above mentioned.

We compare the proposed framework with the faithfulness metric, which is able to improve significantly the classification accuracy but requires maintenance decision-makers to review a much larger number of images. This is evident from its lower precision, indicating that only a small fraction of the flagged samples is effectively anomalous. An exception is observed when applying $\Phi^E$ to the flashover class, which exhibits large standard deviations in both accuracy improvement and in the proportion of images needing review. For example, accuracy improvement is 0% in two out of five folds, while one fold requires reviewing 45% of test images, which is unacceptably high. Overall, the proposed framework outperforms the faithfulness metric in distinguishing between correctly and incorrectly classified image explanations. This is reflected by the $F_1$ score, which is consistently higher for the proposed method.

Table 4: Performance comparisons between the proposed framework and the comparison method based on the faithfulness metric. Mean and standard deviation are calculated across five cross-validation folds; $acc(\Phi)$ indicates the model's initial classification accuracy, whereas $acc(\Psi)$ is the accuracy reclassification by the maintenance operators.

| | | | $acc(\Psi) - acc(\Phi)$ | % of the test images revised by domain expert | Precision | Recall | $F_1$ score |
|---|---|---|---|---|---|---|---|
| EfficientNet | Flashover | Proposed method | +8.46 ±8.06 % | 19.27 ±20.66 | 0.48 ±0.37 | 0.33 ±0.35 | 0.37 ±0.24 |
| | | Comparison method | +9.17 ±17.79 % | 11.57 ±19.22 | 0.58 ±0.29 | 0.17 ±0.29 | 0.35 ±0.37 |
| EfficientNet | Broken | Proposed method | +6.35 ±6.5 % | **11.13 ±8.39** | 0.56 ±0.25 | 0.24 ±0.21 | 0.37 ±0.21 |
| | | Comparison method | +12.87 ±17.64 % | 45.34 ±41.56 | 0.16 ±0.23 | 0.38 ±0.52 | 0.22 ±0.31 |
| MobileNet | Flashover | Proposed method | **+9.91 ±8.54 %** | 14.67 ±11.37 | **0.64 ±0.21** | 0.33 ±0.25 | **0.5 ±0.19** |
| | | Comparison method | +26.18 ±6.69 % | 92.45 ±10.72 | 0.29 ±0.11 | 0.97 ±0.04 | 0.44 ±0.13 |
| MobileNet | Broken | Proposed method | +7.32 ±3.67 % | 15.48 ±9.73 | 0.60 ±0.31 | **0.44 ±0.37** | 0.36 ±0.18 |
| | | Comparison method | +12.05 ±11.06 % | 32.32 ±24.97 | 0.25 ±0.23 | 0.50 ±0.50 | 0.32 ±0.30 |

The presented results utilize Grad-CAM explanations but the proposed framework is versatile with respect to the XAI method employed: Appendix B demonstrates this by presenting results using CartoonX explanations for $\Phi^M$ on the broken class, which lead to similar conclusions.

# 8 Conclusions

In the context of fault diagnosis for large infrastructures through image classification, we propose a novel framework that automatically processes model explanations to identify misclassified images or correctly classified images that rely on shortcuts. Our framework is composed by three modules, which, in principle, are not constraint to specific models. For the considered application, we constrain the selected models to ensure that the framework is post-hoc, model- agnostic with respect to the classification model, which may have already been developed for fault diagnosis of large infrastructures. We demonstrate the proposed framework considering two CNNs as classification models

(MobileNetV3 Small and EfficientNet-B0), leveraging GradCAM to generate explanations and Deep SAD to detect anomalous explanations. The automated processing of explanations significantly reduces the workload of maintenance operators, who are only required to manually review and potentially reclassify images whose explanations deviate from those of correctly classified images.

We applied the proposed framework to a fault diagnostic model tasked with classifying images of power grid insulator shells captured by drones. The results demonstrate that our method outperforms the faithfulness metric by improving classification accuracy while simultaneously reducing the effort required from maintenance operators.

Our framework has used GradCAM and Deep SAD but it is adaptable to any feature attribution XAI method and semi-supervised anomaly detection technique. The objective of the present study has been to present the proposed framework and its use for supporting maintenance; optimization of the performance has not been the focus so that the hyperparameters, including those of the CNN classifiers, Grad CAM, Deep SAD and the threshold, have not been systematically optimized. Further optimizations and evaluations with alternative models can be future research. Specifically, a systematic optimization will require a considering the interactions between the models and the definition of an objective function considering several factors, like the costs of misclassifications, manual classifications and model development. Additionally, it would be interesting to explore the applicability of the proposed framework to other types of data different from images.

Another improvement of the framework could concern the integration of multiple XAI techniques to obtain richer explanations. This could help to mitigate the effects of non-informative explanations, as those provided by CartoonX for images of one class of faults. In fact, since each XAI technique examines model decisions from different perspectives, combining several XAI methods could extend the information on how the DL model is reasoning and potentially improve Deep SAD's (or other semi-supervised AD methods') ability to distinguish anomalous explanations. Additionally, future research will explore strategies to minimize misclassifications and non-causal shortcuts during model training. One potential direction of development is to involve experts in providing explanations and retraining the diagnostic model using a loss function that explicitly incorporates desired explanation characteristics.

# Appendix A. details on hyperparameters setting for Deep SAD

This Section discusses further the setting of the hyperparameters of the Deep SAD models. Table A1 reports the hyperparameters chosen for the four developed combinations of CNN models and assigned fault classes. The Adam optimizer with an initial learning rate of $10^{-3}$ yields good results in three out of four cases but leads to a trivial solution for the broken class with EfficientNet-B0, where all explanations are mapped to the same point. Therefore, other learning rate values have been tested, and a satisfactory solution has been found using a reduced learning rate of $10^{-4}$.

Table A1: Deep SAD Hyperparameters setting used for different CNNs and classes

**EfficientNet-B0 on flashover class**

| | |
|---|---|
| Number of epochs for training the autoencoder = 200 | Autoencoder Early Stopping = 25 epochs |
| Number of epochs for training the Deep SAD = 1000 | Early Stopping = 50 epochs (after epoch 100) |
| Weight decay = $10^{-5}$ | Initial Learning Rate = $10^{-3}$ |
| Batch size = 32 | Weight applied to labeled samples in Eq. 1: $\eta = 10$ |

**EfficientNet-B0 on broken class**

| | |
|---|---|
| Number of epochs for training the autoencoder = 200 | Autoencoder Early Stopping = 25 epochs |
| Number of epochs for training the Deep SAD = 1000 | Early Stopping = 50 epochs (after epoch 100) |
| Weight decay = $10^{-5}$ | Initial Learning Rate = $10^{-4}$ |
| Batch size = 32 | Weight applied to labeled samples in Eq. 1: $\eta = 10$ |

**MobileNetV3 Small on flashover class**

| | |
|---|---|
| Number of epochs for training the autoencoder = 100 | Autoencoder Early Stopping = 20 epochs |
| Number of epochs for training the Deep SAD = 1000 | Early Stopping = 50 epochs (after epoch 100) |
| Weight decay = $10^{-2}$ | Initial Learning Rate = $10^{-3}$ |
| Batch size = 64 | Weight applied to labeled samples in Eq. 1: $\eta = 100$ |

**MobileNetV3 Small on broken class**

| | |
|---|---|
| Number of epochs for training the autoencoder = 200 | Autoencoder Early Stopping = 25 epochs |
| Number of epochs for training the Deep SAD = 1000 | Early Stopping = 50 epochs (after epoch 100) |
| Weight decay = $10^{-2}$ | Initial Learning Rate = $10^{-3}$ |
| Batch size = 64 | Weight applied to labeled samples in Eq. 1: $\eta = 100$ |

Another important hyperparameter to be set is $\eta$ in Eq. 1, which quantifies the importance of the labelled samples in the loss. When the Deep SAD model for the EfficientNet-B0 network is considered, $\eta = 10$ is used to account for the small number of available anomalies. This allows obtaining comparable magnitudes of the terms $\eta \times \mathcal{L}_{LA}$ and $\mathcal{L}_{UL}$ in Eq. 1. However, when using the same value for the MobileNetV3 Small network, the obtained results are not satisfactory as the loss term $\eta \times \mathcal{L}_{LA}$ is too small and, as a consequence, it is ignored during training. For this reason, we have increased the value of $\eta$ to 100, obtaining satisfactory results. Figure A1 shows the evolution of the loss terms during training of the Deep SAD model developed for MobileNetV3 Small and the flashover class. The contributions of the two terms $\mathcal{L}_{UL}$ and $\eta \times \mathcal{L}_{LA}$ from Eq. 1 are comparable. Also, an increase of $\mathcal{L}_{UL}$ corresponds to a decrease of $\eta \times \mathcal{L}_{LA}$, and vice versa, indicating that the model balances the two terms during training to reduce the overall loss.

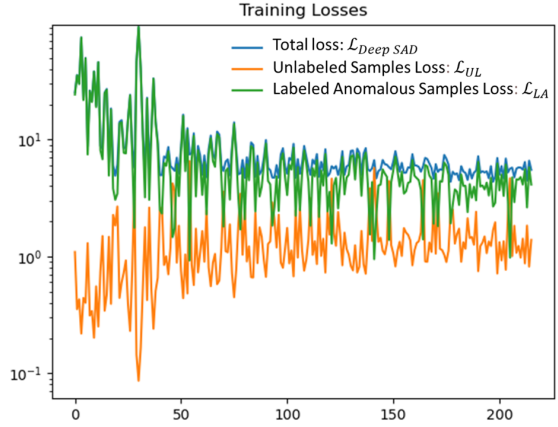

Figure A2: training losses of MobileNetV3 on the flashover class.

## Appendix B. Results obtained using CartoonX

This Section investigates the robustness of the proposed framework to the XAI method used to provide the explanations. Table B1 compares the performance of the proposed framework when CartoonX and GradCAM are used. For ease of comparison, the results previously discussed in Section 7.3 obtained using GradCAM are also reported.

When CartoonX is used, the proposed framework leads to a 6.7% increase in accuracy, with maintenance operators reviewing 19% of the test images. As indicated by the recall value, 40% of misclassifications are identified. Precision, defined as the proportion of reviewed images that effectively need reclassification, amounts at 43%. In comparison, the faithfulness metric leads to a smaller improvement in average accuracy (+4.9%) and requires more maintenance operators' reclassifications (33%). Additionally, it exhibits lower precision (13%) and recall (38%). The superiority of the proposed framework with respect to the application of the faithfulness metric to explanations provided by CartoonX is confirmed by a larger $F_1$ score (0.51 vs 0.25).

The results obtained using CartoonX to gather explanations are comparable to the ones obtained using GradCAM, demonstrating the versatility of the proposed framework, whose output is robust with respect to the choice of the utilized XAI method.

Table B1: comparison between the proposed framework when GradCAM and CartoonX are used as XAI method. The results refer to the faults of the broken class and the case in which MobileNetV3 is used for fault classification. Mean and standard deviation are calculated across five cross-validation folds; $acc(\Phi)$ indicates the model's initial classification accuracy, whereas $acc(\Psi)$ is the accuracy reclassification by the maintenance operators.

|  |  | $acc(\Psi) - acc(\Phi)$ | % of the test images revised by domain expert | Precision | Recall | F1 score |
|---|---|---|---|---|---|---|
| GradCAM | Proposed method | +7.32 ±3.67 % | 15.48 ±9.73 | 0.60 ±0.31 | 0.44 ±0.37 | 0.36 ±0.18 |
| CartoonX | Proposed method | +6.66 ±6.88 % | 20.29 ±25.29 | 0.43 ±0.17 | 0.40 ±0.37 | 0.51 ±0.13 |
| CartoonX | Comparison method | +4.88 ±3.06 % | 33.07 ±8.67 | 0.13 ±0.07 | 0.38 ±0.21 | 0.25 ±0.07 |

The proposed framework failed when applied to the flashover class using explanations obtained by CartoonX. This is attributed to the observed difficulty of CartoonX in providing satisfactory explanations for images of this class. As example, Figure B1 shows one image from flashover class (Figure B1a) and the corresponding explanations obtained using CartoonX and Grad CAM. Figure B1b shows that CartoonX roughly identifies the flashover area, but it struggles in selecting a continuous region and provides a jittery explanation, which is not per-se meaningful and it is difficult to postprocess by the semi-supervised AD module. On the other hand, Figure B1c shows that the explanation by GradCAM highlights a continuous region which includes the flashover area and can be more easily processed by the semi-supervised AD module.

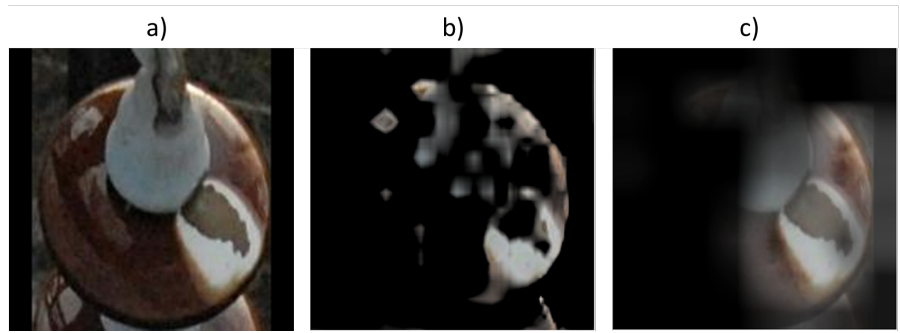

Figure B1: Example of one image from the flashover class (a) and corresponding explanations obtained by CartoonX (b) and Grad CAM (c).